\documentclass[conference]{IEEEtran}
\IEEEoverridecommandlockouts
\usepackage{cite}
\usepackage{amsmath,amssymb,amsfonts}
\usepackage{algorithmic}
\usepackage{graphicx}
\usepackage{textcomp}
\usepackage{xcolor}
\usepackage{booktabs}
\usepackage{cleveref}

\def\BibTeX{{\rm B\kern-.05em{\sc i\kern-.025em b}\kern-.08em
    T\kern-.1667em\lower.7ex\hbox{E}\kern-.125emX}}
\begin{document}

\title{Enhancing Customer Contact Efficiency with Graph Neural Networks in Credit Card Fraud Detection Workflow\\
}



\author{
\IEEEauthorblockN{1\textsuperscript{st} Menghao Huo\textsuperscript{\dag,*}}
\IEEEauthorblockA{\textit{School of Engineering} \\
\textit{Santa Clara University} \\
Santa Clara, CA, USA \\
menghao.huo@alumni.scu.edu \\
ORCID: 0009-0000-0076-5343}

\vspace{+1em}

\IEEEauthorblockN{2\textsuperscript{nd} Qiang Zhu}
\IEEEauthorblockA{\textit{Department of Mechanical and Aerospace Engineering} \\
\textit{University of Houston} \\
Houston, TX, USA \\
qzhu11@uh.edu \\
ORCID: 0009-0002-0981-0635}

\and

\IEEEauthorblockN{1\textsuperscript{st} Kuan Lu\textsuperscript{\dag}}
\IEEEauthorblockA{\textit{School of Electrical and Computer Engineering} \\
\textit{Cornell University} \\
Ithaca, NY, USA \\
kl649@cornell.edu \\
ORCID: 0009-0003-5744-9247}

\vspace{+1em}

\IEEEauthorblockN{3\textsuperscript{rd} Zhenrui Chen}
\IEEEauthorblockA{\textit{Fu Foundation School of Engineering and Applied Science} \\
\textit{Columbia University in the City of New York} \\
New York, NY, USA \\
zc2569@columbia.edu \\
ORCID: 0009-0007-6909-3638}

\thanks{\textsuperscript{\dag}Menghao Huo and Kuan Lu contributed equally to this work.}
\thanks{*Corresponding author: Menghao Huo (menghao.huo@alumni.scu.edu)}
}

\maketitle

\begin{abstract}
Credit card fraud has been a persistent issue since the last century, causing significant financial losses to the industry. The most effective way to prevent fraud is by contacting customers to verify suspicious transactions. However, while these systems are designed to detect fraudulent activity, they often mistakenly flag legitimate transactions, leading to unnecessary declines that disrupt the user experience and erode customer trust. Frequent false positives can frustrate customers, resulting in dissatisfaction, increased complaints, and a diminished sense of security. To address these limitations, we propose a fraud detection framework incorporating Relational Graph Convolutional Networks (RGCN) to enhance the accuracy and efficiency of identifying fraudulent transactions. By leveraging the relational structure of transaction data, our model reduces the need for direct customer confirmation while maintaining high detection performance. Our experiments are conducted using the IBM credit card transaction dataset to evaluate the effectiveness of this approach.
\end{abstract}


\section{Introduction}
Fraud has been a significant challenge for the banking and financial industry since the last century. According to newly released data from the Federal Trade Commission, consumers reported losing over \$10 billion to fraud in 2023, marking the first time fraud losses have surpassed this milestone\cite{govdata2024}. Addressing financial fraud is critical to ensuring sustainable economic growth.

One prominent area of research in combating financial fraud is credit card fraud detection and prevention. Credit card fraud refers to the unauthorized use of funds in transactions, often facilitated through credit or debit cards. In 2023 alone, credit card fraud resulted in losses totaling \$246 million across 114,000 reported cases, representing the highest number of reports among the ten major fraud categories. Notably, the volume of suspicious transactions far exceeds the number of reported fraud cases, underscoring the urgency of improving detection and prevention systems.

\subsection{Fraud Detection and Processing Problem}
Our current bank detection and processing systems heavily rely on fraud detection rules provided by associations such as FICO and contact channels to identify potentially fraudulent transactions. While these systems are designed to safeguard against fraud, they often lead to the incorrect flagging of legitimate transactions as fraudulent. This issue not only disrupts the user experience by causing unnecessary transaction declines but also severely undermines customer trust in financial institutions. When customers face frequent false positives, they may become frustrated with the system, leading to dissatisfaction, increased complaints, and a sense of insecurity. As a result, financial institutions must allocate additional resources to handle these complaints, further increasing operational costs. Over time, this cycle can diminish the overall effectiveness of the fraud detection system, ultimately compromising its reliability and efficiency. Moreover, the loss of trust from customers can have long-term negative effects on the institution's reputation, potentially resulting in reduced customer retention and a decline in new customer acquisition. Therefore, addressing the problem of false positives is critical to ensuring that fraud detection systems remain both effective and customer-friendly.

\subsection{Challenge}
Fraud detection and processing systems face significant challenges that impact their accuracy and efficiency. One key issue is data imbalance, where fraudulent transactions constitute only a small fraction of the total transaction volume. This imbalance often causes models to favor legitimate transactions, leading to missed detections of fraudulent activities. Additionally, feature complexity presents a major hurdle, as fraud detection requires the integration and analysis of diverse data points, such as transaction history, customer behavior, device information, and network activity. This complexity increases computational demands and can slow down detection processes, hindering real-time responses. Furthermore, the scarcity of labeled data exacerbates these difficulties, as machine learning-based systems depend on labeled datasets for training. Fraudulent transactions are rare and challenging to label accurately, and manual labeling is both time-consuming and prone to error. Overcoming these interconnected challenges is essential for building robust and reliable fraud detection systems capable of effectively addressing evolving threats.

Recent progress in data-driven modeling and learning-based systems has significantly enhanced our ability to address complex problems across domains. For example, nearest neighbor methods have seen renewed interest, both in density estimation~\cite{zhao2022analysis} and in reinforcement learning settings with minimax guarantees~\cite{10816316}. In the context of retrieval-augmented generation, RAG-Instruct~\cite{liu2024raginstructboostingllmsdiverse} explores how diverse prompts can guide large language models (LLMs) more effectively. Similarly, BlendSQL~\cite{glenn-etal-2024-blendsql} offers a unified structured language to support hybrid question answering over structured and unstructured data, addressing the limitations of standard prompt-based paradigms.

More recent work has emphasized the integration of LLMs in graph reasoning~\cite{shi2025deepsemanticgraphlearning}, noise-robust multimodal pretraining~\cite{tao2024nevlpnoiserobustframeworkefficient}, and structured thought modeling~\cite{zhang2025ratt}, expanding the boundaries of explainable and generalizable AI. In computer vision, methods like CPDR~\cite{Li_2024_BMVC}, DDUNet~\cite{li2025ddunet}, and causal model-driven super-resolution frameworks such as CausalSR~\cite{lu2025causalsrstructuralcausalmodeldriven} have advanced efficient visual representation learning. Additionally, innovative sensing approaches, such as heartbeat monitoring via RFID and latent diffusion models~\cite{wang2025fine}, demonstrate the growing intersection of signal modeling and generative AI.

\section{Related Work}
\subsection{Bayesian Network Model}

\begin{figure}[h]
  \centering
   \includegraphics[width=0.8\linewidth]{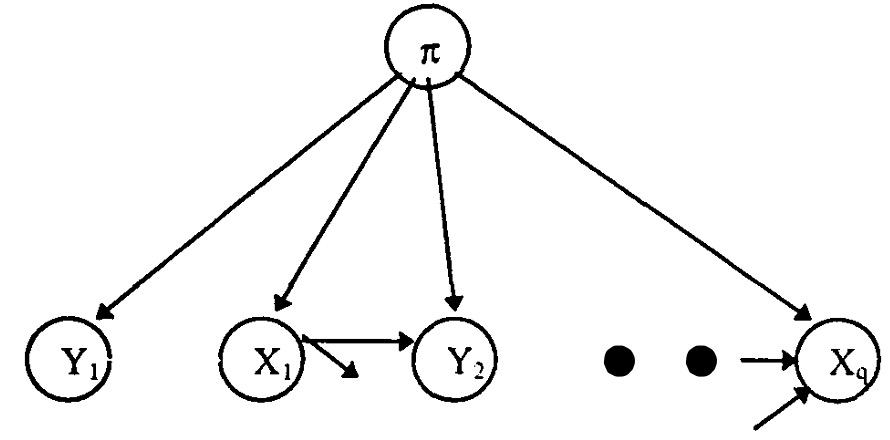}

   \caption{Bayesian Model Network Graph}
   \label{fig:Bayesian}
\end{figure}

Bayesian networks excel at handling uncertainty and predicting rare events~\cite{WANG2024100522}, both critical challenges in fraud detection. By utilizing probabilistic reasoning, they can evaluate the likelihood of fraud based on available evidence, even when data is incomplete or missing, as illustrated in Fig.~\ref{fig:Bayesian}. Additionally, Bayesian networks are particularly effective in identifying rare fraud cases by incorporating prior knowledge and reasoning under uncertainty, enabling accurate predictions even when the available data is sparse or limited~\cite{cite1}. 

However, Bayesian networks face several challenges when applied to large-scale fraud detection. For datasets with billions of transactions, they may struggle to scale efficiently compared to other machine learning techniques like neural networks or tree-based models. Additionally, Bayesian networks rely on the assumption of conditional independence between variables, which may not hold in fraud detection scenarios where complex interactions exist. This, coupled with their computational complexity, can make Bayesian networks resource-intensive and limit their ability to perform in real-time, especially in high-transaction environments with many variables or intricate interdependencies.

In conclusion, Bayesian network models provide a strong, interpretable, and probabilistic approach for fraud detection, particularly for rare event prediction and handling uncertainty. However, their computational cost, scalability issues, and reliance on domain knowledge may pose challenges in large-scale or real-time fraud detection systems.

\subsection{Recurrent Neural Networks}

Recurrent Neural Networks (RNNs) are particularly effective in fraud detection due to their ability to handle sequential data and capture temporal dependencies, making them ideal for modeling the evolving patterns of fraudulent activities over time. RNNs can process raw, time-ordered transaction data without the need for extensive feature engineering, simplifying data preprocessing. Their dynamic adaptability to changes in user behavior allows them to detect deviations from normal patterns, which is crucial for identifying fraudulent transactions. Additionally, RNNs can work in real-time, analyzing ongoing transactions and providing immediate fraud detection insights.

However, RNNs also face several challenges. They are prone to the vanishing/exploding gradient problem, which makes it difficult to capture long-term dependencies in transaction sequences. RNNs can also struggle with irregular or sparse sequences, such as those from users with infrequent activity, limiting their effectiveness. Furthermore, RNNs are vulnerable to overfitting, especially when there is a significant imbalance between fraudulent and non-fraudulent transactions. This issue requires careful regularization and model tuning to ensure reliable performance.

\subsection{Random Forest, RF}
Random forest is an ensemble learning method composed of multiple decision trees~\cite{random_forest}. It typically combines the results of multiple decision trees through voting or averaging to produce the final prediction. In the field of fraud detection, random forest works by first converting transaction records into features and preprocessing them. Then, multiple decision trees are created, each forming decision rules based on transaction features until the predefined criteria of the random forest are met. Once the model is built, it can be used to predict whether a new transaction is fraudulent.
Random forest improves classification accuracy through ensemble learning and reduces the error of individual decision trees. It is robust to data noise and outliers and can calculate feature importance, making it useful for understanding fraud patterns. Additionally, it allows parallel training of multiple decision trees, improving efficiency. However, it cannot capture relationships between transactions, making it less effective in detecting organized fraud. As a result, it is not well-suited for large-scale fraud detection.

\section{Methods}
TSYS, the transaction processing system, identifies a suspicious transaction based on predefined rules, patterns, or anomalies in the transaction data. These rules could include unusual spending behavior, transactions originating from high-risk locations, rapid consecutive transactions, or amounts exceeding typical thresholds for the account. The detection process may involve real-time monitoring of transactions as they occur, leveraging fraud detection algorithms or machine learning models that flag activities deviating from normal account behavior. Once a transaction is flagged as suspicious, TSYS immediately raises an alert to notify the associated bank for further investigation and action.

TSYS detects a suspicious transaction and forwards it to the bank for further review. Upon receiving the alert, the bank contacts the association such as FICO to seek guidance on how to proceed. If the association advises processing the transaction directly, the bank processes it without any additional checks. However, if the association recommends contacting the customer, the bank incorporates an additional step by utilizing a fraud detection model.

If the fraud detection model determines that the transaction is fraudulent, it is immediately marked as such. The bank then takes prompt action on the account to prevent any further losses. In this case, the customer is notified about the fraudulent transaction, but their input is not required, which helps to avoid delays and unnecessary involvement.

\begin{figure}[h]
  \centering
   \includegraphics[width=0.8\linewidth]{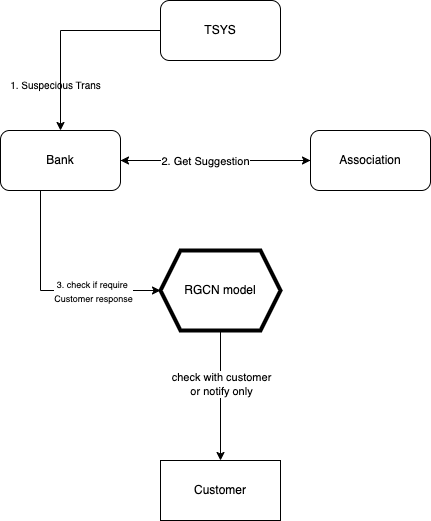}

   \caption{GNN Involved Contact Customer Flow}
   \label{fig:ficoFlow}
\end{figure}

If the fraud detection model identifies the transaction as non-fraudulent, the bank proceeds to contact the customer. In this scenario, the customer is given the opportunity to make the final decision on whether to approve or decline the transaction. This approach effectively minimizes financial losses caused by delays while ensuring that customers are only involved in cases where fraud is not definitively identified. The overview flow is shown in Fig.~\ref{fig:ficoFlow}.

In our model, we use RGCN (Relational Graph Convolutional Network) as the core component. RGCN is a type of Graph Neural Network (GNN) specifically designed for handling heterogeneous graphs, which have different types of nodes and edges. It is capable of learning on multi-relational graphs. In fraud detection scenarios, the data is often heterogeneous. Compared to traditional GCN, RGCN can model different types of relationships more accurately, improving predictive performance.

\subsection{Dataset}
The IBM credit card transaction dataset~\cite{kaggle_creditcard_fraud} is a publicly available resource designed for fraud detection research. Each record contains attributes such as transaction amount, card type, and location, along with a label indicating whether the transaction was fraudulent, the details shows at \cref{tab:dataset}. This provides sufficient feature information in each row to train and evaluate models effectively. Notably, the dataset is entirely synthetic and does not correspond to real individuals or financial institutions. It comprises 24 million unique transactions, 6,000 merchants, 100,000 cards, and 30,000 fraudulent cases, accounting for 0.1\% of the total transactions. Using this dataset to feed into our GNN model, we would work with approximately 16,000 nodes and 100,000 edges, representing the relationships and interactions within the transaction data.

\begin{table}[htbp]
\caption{Dataset Attributes and Types}
\begin{center}
\begin{tabular}{|c|c|}
\hline
\textbf{Column} & \textbf{Dtype} \\
\hline
Year & int64 \\
\hline
Month & int64 \\
\hline
Day & int64 \\
\hline
Amount & float64 \\
\hline
Use Chip & object \\
\hline
Merchant Name & int64 \\
\hline
Merchant City & object \\
\hline
Merchant State & object \\
\hline
Zip & float64 \\
\hline
MCC & int64 \\
\hline
Errors? & object \\
\hline
Is Fraud? & object \\
\hline
card\_id & object \\
\hline
Hour & object \\
\hline
Yours & object \\
\hline
Minute & object \\
\hline
\end{tabular}
\label{tab:dataset}
\end{center}
\end{table}

\subsection{Architecture}

Graph Neural Networks (GNNs) are a type of deep learning model designed to operate on graph-structured data. Unlike traditional neural networks that work with grid-like data such as images or sequences, GNNs can process data represented as nodes (entities) and edges (relationships) in a graph. GNNs learn node, edge, or graph-level representations by aggregating and transforming information from a node's local neighborhood. At each layer, nodes exchange information with their neighbors to update their representations.

\begin{equation}
G = (V, E)
\end{equation}

\begin{figure}[h]
  \centering
   \includegraphics[width=0.8\linewidth]{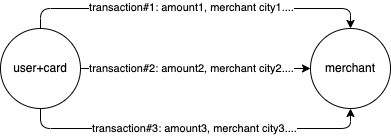}

   \caption{Edges and Nodes in the Graph}
   \label{fig:gnnGraph}
\end{figure}

In this GNN, each node is characterized by two types of attributes: a concatenated string combining the card number and user name, and the merchant name represented as a string. For the edges, we primarily incorporate the attributes of each transaction as edge features, as shown in Fig.~\ref{fig:gnnGraph}. This approach ensures that all relevant information is effectively captured within the graph.

\begin{figure}[h]
  \centering
   \includegraphics[width=0.8\linewidth]{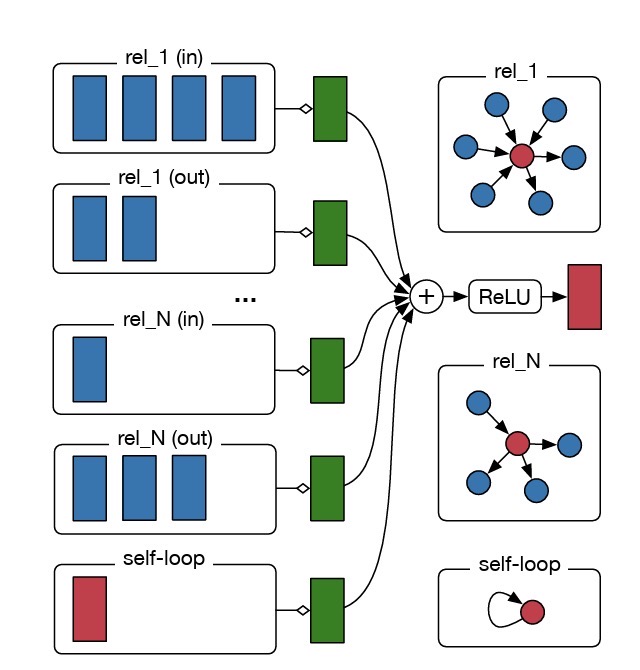}

   \caption{The RGCN model stucture~\cite{rgcnGraph}}
   \label{fig:rgcn}
\end{figure}

In our model, we use RGCN (Relational Graph Convolutional Network) as the core component. RGCN is a type of GNN specifically designed for handling heterogeneous graphs, which have different types of nodes and edges, as illustrated in Fig.~\ref{fig:rgcn}. It is capable of learning on multi-relational graphs. In fraud detection scenarios, the data is often heterogeneous. Compared to traditional GCN, RGCN can model different types of relationships more accurately, improving predictive performance.

The update rule for Relational Graph Convolutional Networks (R-GCN) is given by:

\begin{equation}
h_i^{(l+1)} = \sigma \left( \sum_{r \in R} \sum_{j \in \mathcal{N}_i^r} \frac{1}{c_{ij}^r} W_r^{(l)} h_j^{(l)} + b_r^{(l)} \right)
\end{equation}

In this equation, \( h_i^{(l+1)} \) represents the updated feature vector of node \( i \) at layer \( l+1 \). The term \( h_j^{(l)} \) is the feature vector of a neighboring node \( j \) at the current layer \( l \). The set \( \mathcal{N}_i^r \) consists of the neighbors of node \( i \) that are connected through relation \( r \). The normalization constant \( c_{ij}^r \) helps adjust for the number of neighbors for relation \( r \), typically used to prevent the gradient explosion issue. \( W_r^{(l)} \) denotes the weight matrix specific to relation \( r \) at layer \( l \), and \( b_r^{(l)} \) is the bias term corresponding to relation \( r \) at the same layer. Finally, \( \sigma \) is a non-linear activation function, such as ReLU, which introduces non-linearity and allows the network to learn more complex patterns.

\section{Experiments}
The experiment aimed to create a reliable classifier model that can identify transactions needing customer contact and those that do not. The dataset was split into training and testing sets with an 80:20 ratio. Because the dataset is highly imbalanced, focal loss was used to give more weight to the minority class and reduce the influence of easy-to-classify samples, preventing bias. The model was trained with parameters based on the dataset. The node and edge dimensions were derived from the graph’s features. ReLU was used as the activation function. This function helps the model by adding non-linearity and allowing it to learn more complex patterns. Adam was chosen as the optimization algorithm. It adjusts the model’s parameters and improves performance. Adam combines momentum and adaptive learning rates, making training more efficient and stable.

\begin{figure}[h]
    \centering
    \includegraphics[width=1\linewidth]{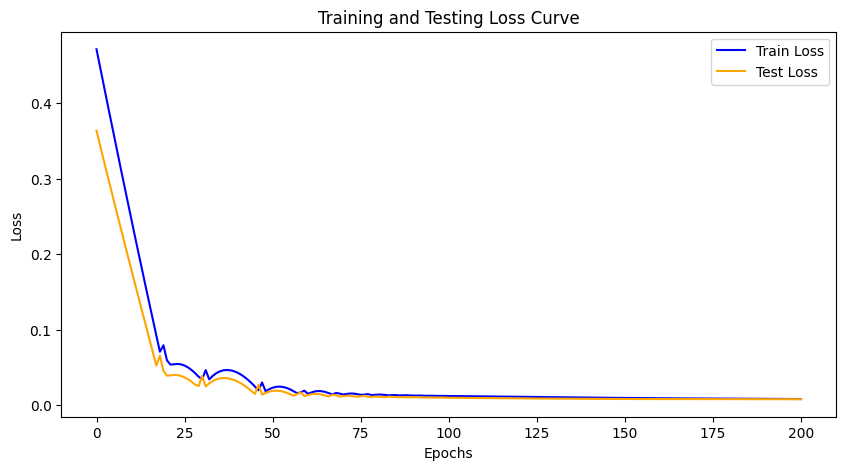}
    \caption{Training and testing loss curve}
    \label{fig:loss-curve}
\end{figure}

The loss curve shown as Fig.~\ref{fig:loss-curve}. The training and testing loss curves in the figure show a sharp decline early on, indicating that the model is learning quickly. Initially, both the training and testing losses are relatively high, but they drop rapidly during the first few epochs. After around epoch 30, the losses stabilize, and both curves approach near-zero values. The training loss (blue curve) decreases more sharply and levels off, suggesting that the model is effectively learning the patterns in the data. The testing loss (orange curve) follows a similar trend but experiences more minor fluctuations, which is typical due to the nature of the testing set. Overall, the steady decrease in loss values across both training and testing phases demonstrates that the model is converging well within the first 200 epochs, even though the testing loss shows slight variability. The final accuracy can achieve 0.9988\%, which is highly approach 100\%.

\section{Conclusion}
In this study, we explored the use of Relational Graph Convolutional Networks (RGCN) for fraud detection and customer verification in credit card transactions. Our research highlights the challenges posed by traditional fraud detection systems, particularly their reliance on predefined rules and customer contact, which often lead to false positives and inefficiencies. By integrating RGCN, our proposed workflow enhances fraud detection accuracy while reducing dependence on direct customer verification. Experimental results demonstrate that the model effectively learns transaction patterns, as indicated by the stable loss curve and near-optimal accuracy of 99.88\%. These findings suggest that RGCN-based fraud detection systems offer a promising solution for financial institutions seeking to improve fraud prevention while minimizing disruptions to legitimate transactions. Future work will focus on further refining the model, optimizing feature engineering, and addressing real-time scalability challenges to enhance its applicability in large-scale financial environments.


\bibliographystyle{IEEEtran}
\bibliography{egbib}

\end{document}